\documentclass[letterpaper]{article} % DO NOT CHANGE THIS
\usepackage{aaai25}  % DO NOT CHANGE THIS
\usepackage{times}  % DO NOT CHANGE THIS
\usepackage{helvet}  % DO NOT CHANGE THIS
\usepackage{courier}  % DO NOT CHANGE THIS
\usepackage[hyphens]{url}  % DO NOT CHANGE THIS
\usepackage{graphicx} % DO NOT CHANGE THIS
\usepackage{natbib}  % DO NOT CHANGE THIS AND DO NOT ADD ANY OPTIONS TO IT
\usepackage{caption} % DO NOT CHANGE THIS AND DO NOT ADD ANY OPTIONS TO IT
\usepackage{algorithm}
\usepackage{algorithmic}

\usepackage{newfloat}
\usepackage{listings}
\DeclareCaptionStyle{ruled}{labelfont=normalfont,labelsep=colon,strut=off} % DO NOT CHANGE THIS
\floatstyle{ruled}
\newfloat{listing}{tb}{lst}{}
\floatname{listing}{Listing}
\usepackage{multirow}
\usepackage{booktabs}
\usepackage{amssymb}

\title{Adapted-MoE: Mixture of Experts with Test-Time Adaption for Anomaly Detection}
\author{
    Tianwu Lei\textsuperscript{\rm 1}\equalcontrib, 
    Silin Chen\textsuperscript{\rm 1}\equalcontrib,
    Bohan Wang\textsuperscript{\rm 1},
    Zhengkai Jiang\textsuperscript{\rm 3},
    Ningmu Zou\textsuperscript{\rm 1, \rm2}\thanks{Corresponding Author}
}
\affiliations{
    \textsuperscript{\rm 1}School of Integrated Circuits, Nanjing University, Suzhou, China\\
    \textsuperscript{\rm 2}Interdisciplinary Research Center for Future Intelligent Chips (Chip-X), Nanjing University, Suzhou, China \\
    \textsuperscript{\rm 3}Hong Kong University of Science and Technology, China \\
    {tianwulei@smail.nju.edu.cn, silin.chen@smail.nju.edu.cn, bohanwang@smail.nju.edu.cn, kaikaijiang.jzk@gmail.com, nzou@nju.edu.cn}\\
}

\usepackage{bibentry}
\begin{document}

\maketitle

\begin{abstract}
    Most unsupervised anomaly detection methods based on representations of normal samples to distinguish anomalies have recently made remarkable progress. However, existing methods only learn a single decision boundary for distinguishing the samples within the training dataset, neglecting the variation in feature distribution for normal samples even in the same category in the real world. Furthermore, it was not considered that a distribution bias still exists between the test set and the train set. Therefore, we propose an Adapted-MoE which contains a routing network and a series of expert models to handle multiple distributions of same-category samples by divide and conquer. Specifically, we propose a routing network based on representation learning to route same-category samples into the subclasses feature space. Then, a series of expert models are utilized to learn the representation of various normal samples and construct several independent decision boundaries. We propose the test-time adaption to eliminate the bias between the unseen test sample representation and the feature distribution learned by the expert model. Our experiments are conducted on a dataset that provides multiple subclasses from three categories, namely Texture AD benchmark. The Adapted-MoE significantly improves the performance of the baseline model, achieving 2.18\%-7.20\% and 1.57\%-16.30\% increase in I-AUROC and P-AUROC, which outperforms the current state-of-the-art methods. Our code is available at \url{https://github.com/}.

\end{abstract}


\begin{figure}[t]
\centering
\includegraphics[width=0.45\textwidth]{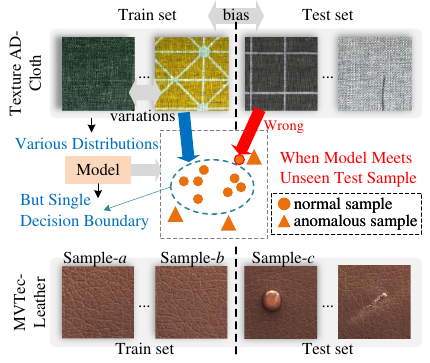} 
\caption{Existing methods construct single decision boundary by learning representations of normal samples, ignoring variations in the feature distribution of samples within the same category as shown in the Texture AD-Cloth \cite{texturead}. Moreover, the test dataset still has a massive distribution of unseen samples. Existing datasets (e.g., MVTec AD dataset \cite{bergmann2019mvtec}) in which similar samples are all in the same distribution are illustrated by Sample-$a$, Sample-$b$, and Sample-$c$.}
\label{figure:1}
\end{figure}

\section{Introduction}

    Anomaly detection recognizes anomalous images and detects anomalous regions, which is a essential method in industrial quality applications \cite{bergmann2019mvtec, liu2024deep}. Because obtaining and labeling anomalous samples is difficult in the real world, unsupervised anomaly detection (UAD) which discriminates outliers by learning normal sample features has gradually become the focus of research \cite{wu2024unsupervised,heckler2023exploring,liu2022unsupervised}. Motivated by the fact that normal samples are easy to collect, many methods learn the features distribution of normal samples by reconstructing them recently \cite{ristea2022self,zhang2023diffusionad}. These methods assume that the reconstruction network can distinguish between representations of anomalous samples based on distributions learned from normal samples, thereby establishing the decision boundary. Other methods are based on synthetic anomalous images which are normal thus learning discriminative image features by deep learning models\cite{zavrtanik2021draem}. These methods intensely depend on the quality of the synthetic anomaly images as well as on more empirical knowledge about the defect patterns. Some methods also use memory-bank \cite{park2020learning,wang2023multimodal,hu2024dmad} to store features of normal samples and discriminate anomalous samples by calculating feature similarity. These methods ignore the existence of unseen samples within the testing process. We summarise these current methods as shown in Figure \ref{figure:1}, where the methods uniformly learn representations distribution for normal samples and build a single decision boundary in the same category based on the distribution. In the test time, samples outside the decision boundary are considered anomalous samples. 

    The aforementioned methods demonstrate optimal performance due to the consistency in the training datasets and exhibit minimal distribution bias between the train set and test set (e.g. Sample-$a$, Sample-$b$, and Sample-$c$ from MVTec in Figure \ref{figure:1}). However, the real samples are affected by variations in the lighting conditions, equipment, camera position, and other factors during the acquisition process. It results in a variation in the distribution of the samples used to learn the representations, as well as the samples to be detected. As shown in Figure \ref{figure:1}, practical applications suffer from a large number of samples in the same category that are still ``novel type'' (e.g. different color, material in Texture AD-Cloth) exacerbating the variation in the train set. Furthermore, it is possible that the test data and the training data, which belong to the same category, may exhibit distribution bias. Unseen normal samples may be projected outside the single decision boundary, potentially leading to significant inaccuracies. In this paper, we formulate the mentioned issue in terms of two definitions. (1) Various complicated feature distributions exist in the training samples. As shown in Figure \ref{figure:1}, samples in Texture AD-Cloth are collected from the same category (cloth), but each sample is in a completely independent data distribution due to color and material differences. It indicates that a single decision boundary in the training process is not sufficient to distinguish all samples of the same category. (2) Distribution bias in the test set and train set for normal and anomalous samples. As shown in Figure \ref{figure:1}, the test set samples are unseen compared to the training set in Texture AD-Cloth. The application of the decision boundary derived from the training samples has been demonstrated to result in inconsistencies and inaccuracies.

    As a significant number of real samples are excluded within the dataset, we propose a new method called Adapted Mixture of Experts (Adapted-MoE) to solve the above issue. Firstly, we use a pre-trained model on ImageNet, similar to \cite{roth2022towards, liu2023simplenet,lei2023pyramidflow}, to extract feature embeddings based on the training dataset. To address the distribution of normal samples with various independent patterns, we introduced a Mixture of Expert models to deconstruct distinct distributions over the feature embeddings. A representation learning based Routing Network is proposed to route feature embeddings to expert models dedicated to discrimination. The proposed mixture of experts can learn multiple independent distributions of various normal subclasses and model several decision boundaries, eliminating the negative impact of constructing a single distribution for samples in the same category. In the testing process, we propose a Test-Time Adaption to calibrate with the distribution of unseen sample representation. Specifically, we assume that a random normal sample has a distribution with a certain pattern in the feature space. We leverage the mean and variance of the normal samples to unify the feature embeddings under the same distribution as the learned specific pattern before inputting them into the expert model via a normalization method. The major contributions of this paper are summarized as follows:
    \begin{itemize}
        \item To our knowledge, our proposed Adapted-MoE firstly investigates the challenging problem of variation in the train set and bias between the train set and test set for anomaly detection. 
        \item We propose a MoE model for learning normal sample feature distribution for different subclasses. Moreover, we also designed a routing network based on representation learning to distinguish normal samples. A simple and effective test-time adaption is proposed to solve the unseen sample bias in the testing process.
        \item We conduct extensive experiments to confirm the effectiveness of the Adapted-MoE on the new benchmark, called the Texture AD benchmark. This benchmark aggregates multiple samples of different patterns (e.g. different colors, different conditions of imaging) within the same category, which is much closer to the reality of the situation. The experimental results show that the proposed method significantly outperforms the previous state-of-the-art.
        % (加入算力精度对比)
    \end{itemize}

\begin{figure*}[t]
\centering
\includegraphics[width=0.85\textwidth]{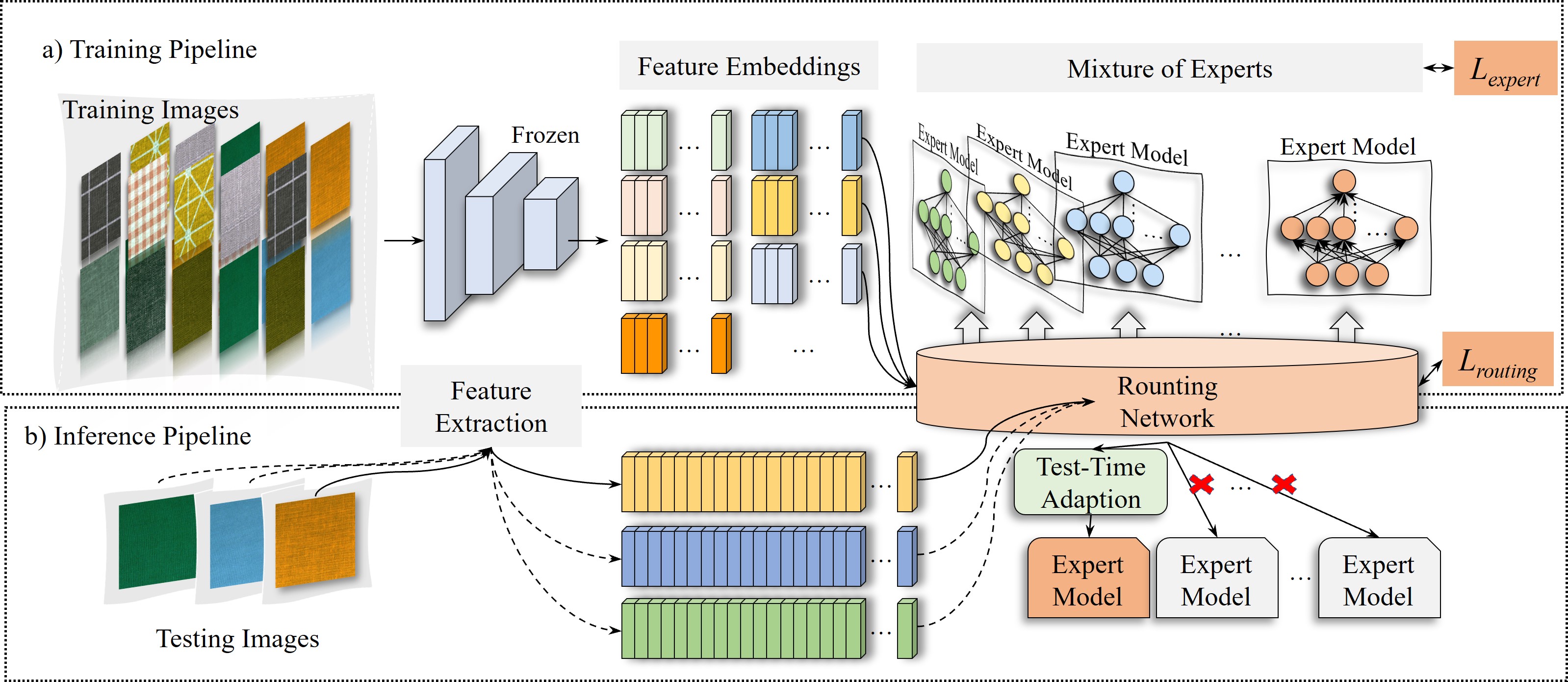} 
\caption{Overview of Adapted-MoE. First a frozen backbone is employed to conduct feature extraction on the samples. Subsequently, the extracted feature embeddings are divided into different expert models for training through a routing network, where the training loss consists of the routing loss $L_{routing}$ and the loss of the expert model $L_{expert}$. In the testing phase, Test-Time Adaption calibrates the routed features to eliminate distribution bias before anomaly detection.}
\label{figure:2}
\end{figure*}

\begin{figure}[t]
\centering
\includegraphics[width=0.45\textwidth]{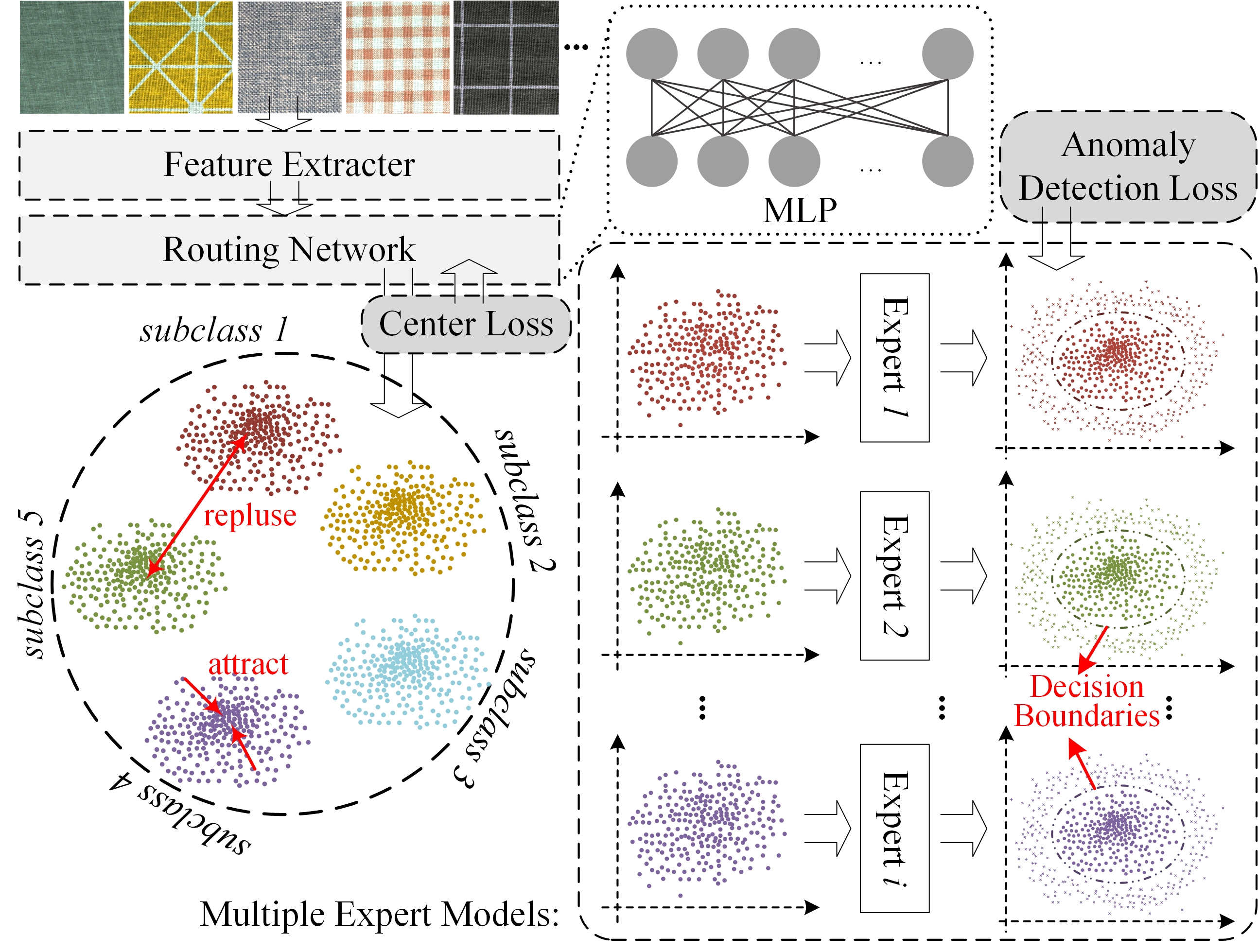} 
\caption{Mixture of Experts. For a mini-batch of feature embeddings, the center loss is utilized in the routing network to divide them into different subclasses during the training process. Simple expert models construct multiple decision boundaries in independent feature spaces for different subclasses. }
\label{figure:3}
\end{figure}

\section{Related Works}
    With the rapid development of deep learning, most anomaly detection methods are divided into reconstructed-based and anomalous simulation-based models \cite{li2023survey,liu2024deep}. \\
    \textbf{Reconstruction-based approach.} The reconstruction-based approaches assumed that anomalous samples cannot be correctly reconstructed by a feature learning method constructed based on normal samples\cite{cao2024survey}. Early reconstructed-based methods used Auto-encoder network to construct the decision boundary by learning the low-dimensional features of normal images to obtain latent variables and reconstruct the normal samples using a decoder \cite{bergmann2019improving,baur2019deep,mishra2021vt}. As generative models have developed, some approaches have utilized generative adversarial networks (GANs)\cite{goodfellow2014generative} to improve the quality of reconstruction \cite{schlegl2017unsupervised,akccay2019skip,liang2023omni}. Owing to the training instability of GANs, some methods combining auto-encoder networks and GANs have been proposed to better model normal samples \cite{zhou2020encoding,contreras2023generative}. Recently, diffusion models based methods have been widely used in anomaly detection tasks with their powerful generative ability \cite{mousakhan2023anomaly,wu2024unsupervised,dai2024generating}. Reconstruction-based methods rely exclusively on normal samples already present within the training set, ignoring the features of samples outside the training set. Therefore, the performance of such methods is greatly limited by the data quality of the normal samples as well as the learning ability of the reconstruction network. \\
    \textbf{Synthesizing-based approach.}The synthesizing-based approach considered the anomalous as noise, and after adding the synthesized defects to the normal samples, the network model was trained to recover them as the corresponding original images \cite{lin2024comprehensive,duan2023few,zhang2023destseg}. It was an intuitive way to add random Gaussian noise to normal samples \cite{sabokrou2018adversarially,haselmann2018anomaly}. However, random noise cannot accurately synthesize real-world anomalous patterns. The distribution of anomalous could be represented even more based on a well-designed mask \cite{yan2021learning,zavrtanik2021reconstruction}. Recently, several feature-based methods have been developed to fit the real anomalous distribution by generating anomalous features embedding \cite{liu2023simplenet,cao2023anomaly,yang2023slsg}. Such methods depend on empirical prior knowledge to construct defective patterns and are therefore difficult to generalize widely to the real world.

\section{Method}
The proposed Adapted MoE is elaborately introduced in this section. As shown in Figure \ref{figure:2}, Adapted-MoE consists of a feature extractor, a routing network, test-time adaption, and several expert models. Specifically, We adopt fixed pre-trained CNNs on ImageNet\cite{deng2009imagenet} as the feature extractor. The features from several stages are collected. Then these features are resized to the same size and concatenated across channel dimensions to restructure the feature maps. Subsequently, the expert model with the highest correlation is assigned via the routing network. The test-time adaptation method is then employed to transfer the feature to the space that can be handled by the selected expert model. Finally, anomaly detection is achieved by the expert model.

\subsection{Mixture of Experts}
Most anomaly detection methods construct the feature space based on normal samples. However, the feature distribution of normal samples in the same category is still diverse, and a single decision boundary will lead to an inaccurately determined outcome. Therefore, we propose a mixture expert model to divide normal samples from the same category into multiple expert models to learn the different feature distributions of multiple subclasses during the training process. Firstly, given the $i_{th}$ training sample's feature maps ${\rm X_i} \in {\mathbb{R}^{C \times H \times W}}$ where $C$, $H$ and $W$ represent the channels, height and width of feature maps, feature embedding $x_i$ are firstly obtained by a projection layer and a global average pooling(GAP) layer $f^{GAP}$. The projection layer is composed of a  $3\times3$ convolution $\mathbf{W}_i$, $\mathbf{b}_i$ to projection feature maps from ImageNet to the anomaly detection feature space:
\begin{equation}
    x_i = {f^{GAP}}(\sigma (\mathbf{W}_iX_i + \mathbf{b}_i)) \in \mathbb{R}^{C}
\end{equation}
Inspired by \cite{wen2016discriminative}, we classify $m$ training samples using a designed center loss in our routing network.
\begin{equation}
 {L_{routing}} =\sum\limits_{i = 0}^m {\alpha \parallel {x_i} - {c_k}{\parallel ^2} - (1 - \alpha ){y_i}\log (\frac{{{e^{{w_i}{x_i}}}}}{{\sum\nolimits_j^n {{e^{{w_j}{x_i}}}} }})} 
\end{equation}
where $m$ represents the $mini-batch$ in the training process, $c_k$ denotes the center of the $k_{th}$ subclass in the traning set and is updated per steps, $y_i$ represents the subclass label for $i_{th}$ normal sample, $n$ is the number of subclass in the training set and also the number of experts, $w\in \mathbb{R}^{C \times n}$ is the classifier matrix and $\alpha$ is weight adjustment parameter, with a value range of $0$ to $1$. As shown in Figure \ref{figure:3}, the samples in the same subclasses are converged to the center of the subclasses by minimizing the above objective function, and the samples from different subclasses will be far away from each other in the feature space. During the inference process, $x_i$ is routed to the expert model with maximize $x_i * c_k$  which denotes the cosine distance between $x_i$ and $c_k$, and the final score will be calculated by the softmax function. 

After obtaining the feature embedding of the subclasses, we simply design a multi-layer perceptron as an expert model to construct decision boundary for independent subclass. We use feature embedding to randomly generate noise vectors and train expert models based on synthetic anomaly detection methods and the loss of expert $L_{expert}$ same as \cite{liu2023simplenet}. The total loss $L_{total}$ is described by: 
\begin{equation}
{L_{total}} = {L_{routing}} + \sum\limits_{i = 1}^m {L_{expert}^i}
\end{equation}
Ultimately, the final anomaly detection score is obtained by aggregating the results of multiple expert models as follows:
\begin{equation}
result = \frac{{\sum\limits_i^k {{w_i}{x_{test}^{i}}Expert({x_{test}^{i}})} }}{{\sum\limits_i^k {{w_i}{x_{test}^{i}}} }}
\end{equation}

\subsubsection{Normalization.}

It is worth emphasizing that due to the similarity of the anomaly detection samples, the feature distribution of the different subclasses that are projected into the feature space is not uniform \cite{reiss2023mean}. Therefore, we adopt the normalization to constrain the value range of the feature embedding $x_i$. It effectively separates the feature of different subclasses so that they can be more evenly distributed in the feature space, which can be expressed as ${x_i} = \frac{{{x_i}}}{{\parallel {x_i}\parallel }}$. As mentioned above, the routing network is scored by cosine similarity and softmax of the classifier matrix $w$ and feature embedding $x_i$. Benefiting from the monotonicity of softmax, the normalized feature embedding does not affect the routing score.

\begin{figure}[t]
\centering
\includegraphics[width=0.45\textwidth]{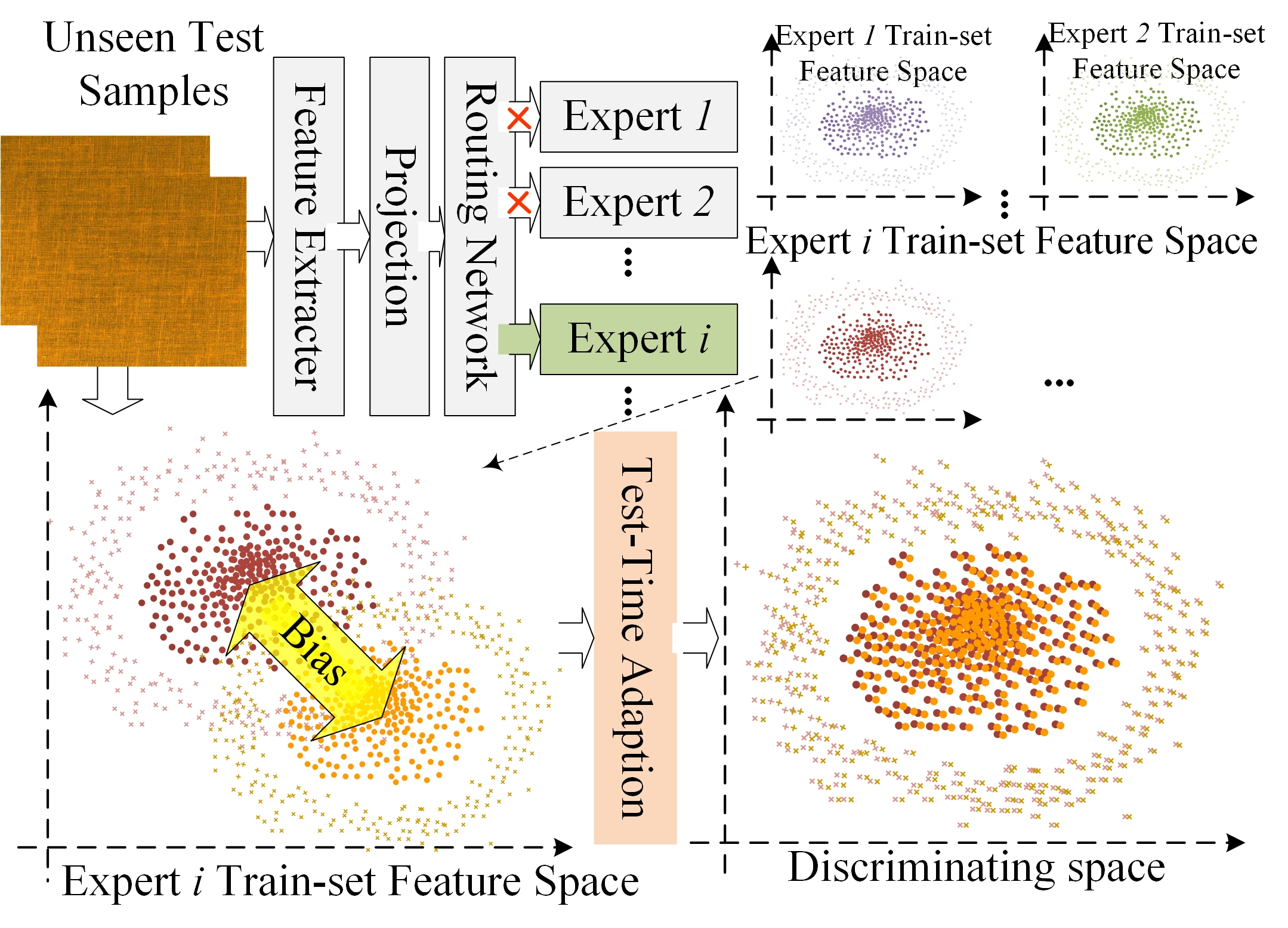} 
\caption{Test-Time Adaptation. Since the test samples do not appear in the training phase, the distribution of the samples at the testing has bias with the distribution of the samples learned by the expert model. We eliminate the distance between the two distributions by Test-Time Adaptation, to unify the position of the decision boundary.}
\label{figure:4}
\end{figure}

\subsection{Test-Time Adaption}
Existing methods construct feature spaces and decision boundaries based on normal samples making unseen samples considered anomalies which makes many out-of-distribution subclass samples misclassified. Based on our proposed MoE, the normal sample in the same category is divided into multiple subclasses routed to different expert models to construct independent decision boundary. However, as shown in Figure \ref{figure:4}, the feature space learned by the expert model based on the existing training set still suffers from a bias in the feature distribution of the unseen subclasses. We assume that the feature distributions of the unseen subclasses have a certain feature distribution in feature space.  Thus their decision boundaries can be obtained by simply eliminating the inconsistency of the feature distributions in the inference process. In this paper, we define this bias as distribution distance and propose a test-time adaptation method to eliminate the bias between unseen samples and training samples.

Firstly, given the feature embedding $x_{test} $ of the test sample, the closest subclass center embedding $c_k$ of the test sample in the feature space can be found by the routing network. As shown in Figure \ref{figure:4}, the test sample distribution and the learned distribution are similar but have a distance gap. Since the $k_{th}$ expert model is based on training data with center $c_k$ and standard deviation which is denoted as $std$ to construct the decision boundary. Therefore, we calibrate the distribution of test embeddings $x_{test}$ to the feature space of the $k_{th}$ expert model to unify the decision boundary:
\begin{equation}
x_{test}^{'} = \frac{{({x_{test}} - mean({x_{test}})) \times std}}{{std({x_{test}})}} + {c_k}
\end{equation}
\begin{equation}
x_{test}^{'} = \frac{{x_{test}^{'}}}{{\parallel x_{test}^{'}\parallel }}
\end{equation}
We use the center of the training data to make the feature distributions with the same measure by mean and variance and subsequently normalize the corrected embeddings $x_{test}^{'}$ to obtain the final decision boundary.

% In the Routing Network, we can know that the data used to train the expert model E has a representation in the feature space that is close to the feature  of the current test data, thereby minimizing\\
% ||F-c||2\\
% Because the binary classification network of the expert model is trained based on data with a center of c and a standard deviation of std, having a representation distribution close to the training data can achieve higher detection accuracy. We transform the test data representation as follows\\
% Ftest=(Ftest-mean(Ftest)*(std/std(Ftest)))+cl\\

\begin{table*}[t]
\caption{Image-AUROC (\%) comparison with the state-of-the-art methods on Texture AD dataset. }
\label{tab:1}
\centering
\begin{tabular}{c|c|cccccc|c}
\toprule[1.5pt]
% \multirow{2}{*}{Category}               & \multirow{2}{*}{subclass}   & \multirow{2}{*}{SimpleNet}  & \multirow{2}{*}{PyramidFlow} & \multirow{2}{*}{DRAEM} & \multirow{2}{*}{Mean-Shift} & \multirow{2}{*}{MSFlow} & \multirow{2}{*}{EfficientAD} & \multirow{2}{*}{Ours} \\ &&&&&&&& \\ \midrule

\multirow{2}{*}{Category}               & \multirow{2}{*}{subclass}   & {SimpleNet}  & {PyramidFlow} & {DRAEM} & {Mean-Shift} & {MSFlow} & {EfficientAD} & \multirow{2}{*}{Ours} \\ &&\textit{(2023)}&\textit{(2023)}&\textit{(2021)}&\textit{(2023)}&\textit{(2024)}&\textit{(2024)}& \\ \midrule

\multirow{6}{*}{Cloth} & \textit{subclass1}  & 65.08      & 57.88       & 57.58 & \textbf{66.22}      & 50.00  & 65.65       & 57.98 \\
                       & \textit{subclass2}  & 59.26      & 63.18       & 50.21 & 33.66      & 54.01  & \textbf{76.98}       & 62.31 \\
                       & \textit{subclass3}  & 58.83      & 60.74       & 55.44 & 66.21      & 50.00  & 55.69       & \textbf{84.61} \\
                       & \textit{subclass4}  & \textbf{70.40}      & 59.39       & 58.01 & 65.69      & 50.00  & 42.38       & 60.77 \\
                       & \textit{subclass5}  & 68.47      & 49.72       & 55.95 & 39.54      & 50.14  & \textbf{72.20}       & 71.96 \\ \cmidrule{2-9}
                       & \textit{Average}    & 64.41     & 58.18       & 55.44 & 54.26      & 50.83  & 62.58       & \textbf{67.53} \\ \midrule
\multirow{5}{*}{Wafer} & \textit{subclass1}  & 52.11      & 55.54       & 55.69 & 52.83      & 51.19  &     50.28   & \textbf{67.30} \\
                       & \textit{subclass2}  & \textbf{59.66}      & 43.35       & 57.09 & 53.29      & 49.78  &     42.25       & 55.95 \\
                       & \textit{subclass3}  & 53.66      & 52.76       & \textbf{59.22} & 55.44      & 53.64  &     50.23       & 51.71 \\
                       & \textit{subclass4}  & 50.68      & 46.36       & 52.46 & 48.28      & 50.00  &     45.51       & \textbf{59.36} \\ \cmidrule{2-9}
                       & \textit{Average}    & 54.03      & 49.50       & 56.12 & 52.47      & 51.15  & 47.07       & \textbf{58.58} \\ \midrule
\multirow{6}{*}{Metal} & \textit{subclass1}  &59.07           & 52.87       & 52.07 &                        44.34      &  62.90&        65.27&\textbf{65.60}      \\
                       & \textit{subclass2}  &59.87           & 48.74       & 56.32 & 47.39      &  53.54&        55.46&\textbf{66.19}      \\
                       & \textit{subclass3}  &57.83           & 58.92       & 51.48 & 45.04      &  59.78&        \textbf{68.73}&66.57      \\ \cmidrule{2-9}
                       & \textit{Average}    &58.92           & 53.51       & 53.29 & 45.59      &  58.74&        63.30&\textbf{66.12}  \\ \bottomrule[1.5pt]
\end{tabular}%
\end{table*}

\begin{table*}[t]
\caption{Pixel-AUROC (\%) comparison with the state-of-the-art methods on Texture AD dataset.}
\label{tab:2}
\centering
\begin{tabular}{c|c|ccccc|c}
\toprule[1.5pt]
% \multirow{2}{*}{Category}               & \multirow{2}{*}{subclass}   & \multirow{2}{*}{SimpleNet}  & \multirow{2}{*}{PyramidFlow} & \multirow{2}{*}{DRAEM} &  \multirow{2}{*}{MSFlow} & \multirow{2}{*}{EfficientAD} & \multirow{2}{*}{Ours} \\ &&&&&&& \\ \midrule

\multirow{2}{*}{Category}               & \multirow{2}{*}{subclass}   & {SimpleNet}  & {PyramidFlow} & {DRAEM}  & {MSFlow} & {EfficientAD} & \multirow{2}{*}{Ours} \\ &&\textit{(2023)}&\textit{(2023)}&\textit{(2021)}&\textit{(2024)}&\textit{(2024)}& \\ \midrule
\multirow{6}{*}{Cloth} & \textit{subclass1}  & 58.30      & {68.00} & {60.99} & {56.11} & {62.76} & \textbf{80.94} \\
                       & \textit{subclass2}  & 51.52     & {57.06} & {65.36} & {63.14} & {58.92} & \textbf{68.41} \\
                       & \textit{subclass3}  & 63.48      & {60.74} & {56.91} & {51.66} & {47.08} & \textbf{74.43} \\
                       & \textit{subclass4}  & 70.68      & {57.26} & {53.45} & {47.44} & {38.75} & \textbf{76.50} \\
                       & \textit{subclass5}  & 54.47      & {34.84} & {77.03} & {42.23} & {61.77} & \textbf{79.95}\\ \cmidrule{2-8}
                       & \textit{Average}    & 59.69      & {55.58} & {62.75} & {52.12} & {53.86} & \textbf{76.05} \\ \midrule
\multirow{6}{*}{Wafer} & \textit{subclass1}  & 57.18      & {51.23}     & 44.91 & 44.91      &{55.76} & \textbf{60.74}            \\
                       & \textit{subclass2}  & \textbf{66.16}      & {39.47}     & 34.10 & 34.10      &{33.98} & 60.81             \\
                       & \textit{subclass3}  & 57.58      & {51.52}     & 35.01 & 35.01      &{51.53} & \textbf{65.40}              \\
                       & \textit{subclass4}  & 53.40      & {44.63}     & 43.59 & 43.59      &{40.02} & \textbf{66.63}             \\ \cmidrule{2-8}
                       & \textit{Average}    & 58.58      & {46.71} & {39.40} & {39.40} & {45.32} & \textbf{63.40} \\ \midrule
\multirow{6}{*}{Metal} & \textit{subclass1}  &62.27     & 53.42       &58.41                                                 &     65.37&        59.69& \textbf{73.98}\\
                       & \textit{subclass2}  &58.33      & 48.86       &51.53                          &     57.34&        51.04& \textbf{69.48}\\
                       & \textit{subclass3}  &58.97      & 57.67       &57.31  
                                             &     60.37&        54.91& \textbf{77.81}\\ \cmidrule{2-8}
                       & \textit{Average}    &59.86      & 53.31       &55.75
                                             &     61.02&        55.21& \textbf{73.76}\\ \bottomrule[1.5pt]
\end{tabular}
\end{table*}

\begin{table*}[t]
\caption{Ablation Study of Structure for Adapted-MoE.}
\label{tab:3}
\centering
\scalebox{1.0}{
\begin{tabular}{ccc|ccc|ccc}
\toprule[1.5pt]
\multirow{2}{*}{+MoE}  & \multirow{2}{*}{+TTA} &\multirow{2}{*}{+Norm}  &\multicolumn{3}{c|}{Average P-AUROC(\%)}  &\multicolumn{3}{c}{Average I-AUROC(\%)} \\ \cmidrule{4-9} 

 & & & Cloth & Wafer & Metal &Cloth & Wafer & Metal \\ \midrule
 -          & -           &-& 59.69 & 58.58 & 59.86 & 64.41&54.03& 58.92\\ \midrule
 \checkmark & -           &-& 53.93(-5.76) & 58.10(-0.48) & 60.27(+0.41) &58.98(-5.43)&54.95(+0.92)&57.71(-1.21)\\
 -          &\checkmark  &-& 65.96(+6.27) & 59.26(+0.68) & \textbf{73.76(+13.9)} &53.05(-11.36)&54.94(+0.91)&\textbf{66.12(+7.20)}\\ 
 \checkmark & -           &\checkmark& 56.88(-2.81) & 56.23(-2.35) &60.56(+0.70) &58.41(-6.00)&52.40(-1.63)&55.41(-3.51)\\
 -          &\checkmark  &\checkmark& 74.36(+14.67) & 59.73(+1.15) & 63.17(+3.31) &62.83(-1.58)&54.52(+0.49)&55.15(-3.77)\\
 \checkmark &\checkmark  &-& 61.45(+1.76) & 54.42(-4.16) & 56.74(-3.12)  &57.41(-7.00)&54.07+0.04)&53.95(-4.97)\\ 
 \checkmark &\checkmark  &\checkmark& \textbf{76.05(+16.3)} &\textbf{60.15(+1.57)} &69.10(+9.24) &\textbf{67.53(+3.12)}&\textbf{56.21(+2.18)}&55.50(-3.42)\\ \bottomrule[1.5pt]
\end{tabular}
}
\end{table*}

\begin{table}[t]
\caption{P-AUROC(\%) / I-AUROC(\%) for Loss Choices.}
\label{tab:4}
\centering
\scalebox{1.0}{
\begin{tabular}{c|ccc}
\toprule[1.5pt]
 Loss& Cloth& Wafer& Metal \\ \midrule
 Softmax & 74.92/55.48  &58.97/53.24  & \textbf{69.14}/54.92\\
 CenterLoss &\textbf{76.05/67.53} &\textbf{60.15/56.21} &69.10/\textbf{55.50} \\
 \bottomrule[1.5pt]
\end{tabular}
}
\end{table}

\section{Experiments}
% \begin{figure}[t]
% \centering
% \includegraphics[width=0.3\textwidth]{AAAI2025/Figs/Figure5.jpg} 
% \caption{Comparison of Texture AD and MVTec.}
% \label{figure:5}
% \end{figure}
\subsection{Datasets and Metrics}
\textbf{Datasets.} As shown in Figure \ref{figure:1}, existing datasets sampling data are similarly distributed in the same category (e.g., MVTec \cite{bergmann2019mvtec}). To validate the proposed Adapted-MoE, we use a new dataset named Texture AD benchmark\cite{texturead} in the experiments. The Texture AD benchmark is an anomaly detection dataset, which contains sampled images and defect annotations for three categories, cloth, metal, and wafer. Significantly, the Texture AD dataset provides multiple different types in subclasses under each category providing samples of various distributions. In the cloth, it provides $15$ different subclasses of cloth to represent different distributions. There are $14$ different wafer types included in the wafer category. In the metal category, $10$ different types of metal are likewise provided to validate the anomaly detection for different distributions. To validate our method, we choose $10$ subclasses for training and $5$ unseen subclasses for testing in the cloth dataset, $4$ unseen subclasses in the wafer dataset and $3$ unseen subclasses in the metal dataset. All images in this dataset are captured using a high-resolution industrial camera (MV-CS200-10 GC) at $5472\times3648$ pixels and cropped to $1024\times1024$. \\
\textbf{Metrics.} For anomaly detection results, we use the Area Under the Receiver Operating Curve (AUROC) to evaluate our proposed model comprehensively same as other works. Image-level anomaly detection performance is measured via the standard AUROC, denoted as I-AUROC. Moreover, a pixel-level AUROC (P-AUROC) is used to evaluate the anomaly localization.
\subsection{Implementation Details}
All experiment codes are implemented based on the Pytorch framework and all the models are trained with one NVIDIA GeForce RTX 4080 ($16$ GB memory) for acceleration. We validated the effectiveness of the Adapted-MOE using SimpleNet\cite{liu2023simplenet} as our baseline. For the baseline, a pre-trained WideResNet50\cite{zagoruyko2016wide} is used already as a feature extractor which is frozen in both training and testing processes. For fair comparisons, the SimpleNet with Adapted-MoE is trained for $160$ epochs with a batch size of $8$ and the learning rate is from $0.0001$ to $0.0002$. In Gaussian noise $N(0, \sigma^2)$, $\sigma$ is set by default to $0.015$. All experimental results are the mean of $3$ replicates.
\subsection{Comparisons with State-Of-The-Arts}
We compare the proposed Adapted-MoE with a number of state-of-the-art approaches on Texture AD benchmark, including SimpleNet\cite{liu2023simplenet}, EfficientAD\cite{batzner2024efficientad}, PyramidFlow\cite{lei2023pyramidflow}, DREAM\cite{zavrtanik2021draem}, Mean-shifted \cite{reiss2023mean} and MSFlow\cite{zhou2024msflow}. Firstly, we compared the performance of anomaly detection. Since current methods lack consideration of unseen subclasses for testing, our algorithm demonstrates superior performance. As shown in Table \ref{tab:1}, excellent results are achieved by our Adapted-MoE on most of the unseen subclasses in three categories. Moreover, our proposed method outperforms other methods in average I-AUROC accuracy on the test set of cloth, wafer, and metal by $67.53\%$, $58.58\%$, and $66.12\%$, respectively. To further demonstrate the excellence of our method, we secondly compare the capability of anomaly localization on novel unseen data.  As shown in Table \ref{tab:2}, we compare the values of P-AUROC with state-of-the-art methods on three categories in Texture AD. The results show that our method outperforms existing methods in unseen subclass performance for each category as well as average accuracy. The average P-AUROC of our proposed method is $76.05\%$, $63.40\%$, and $73.76\%$ for cloth, wafer, and metal. The results of visualization compared with SOTA are detailed in the Appendix.

\begin{figure}[t]
\centering
\includegraphics[width=0.4\textwidth]{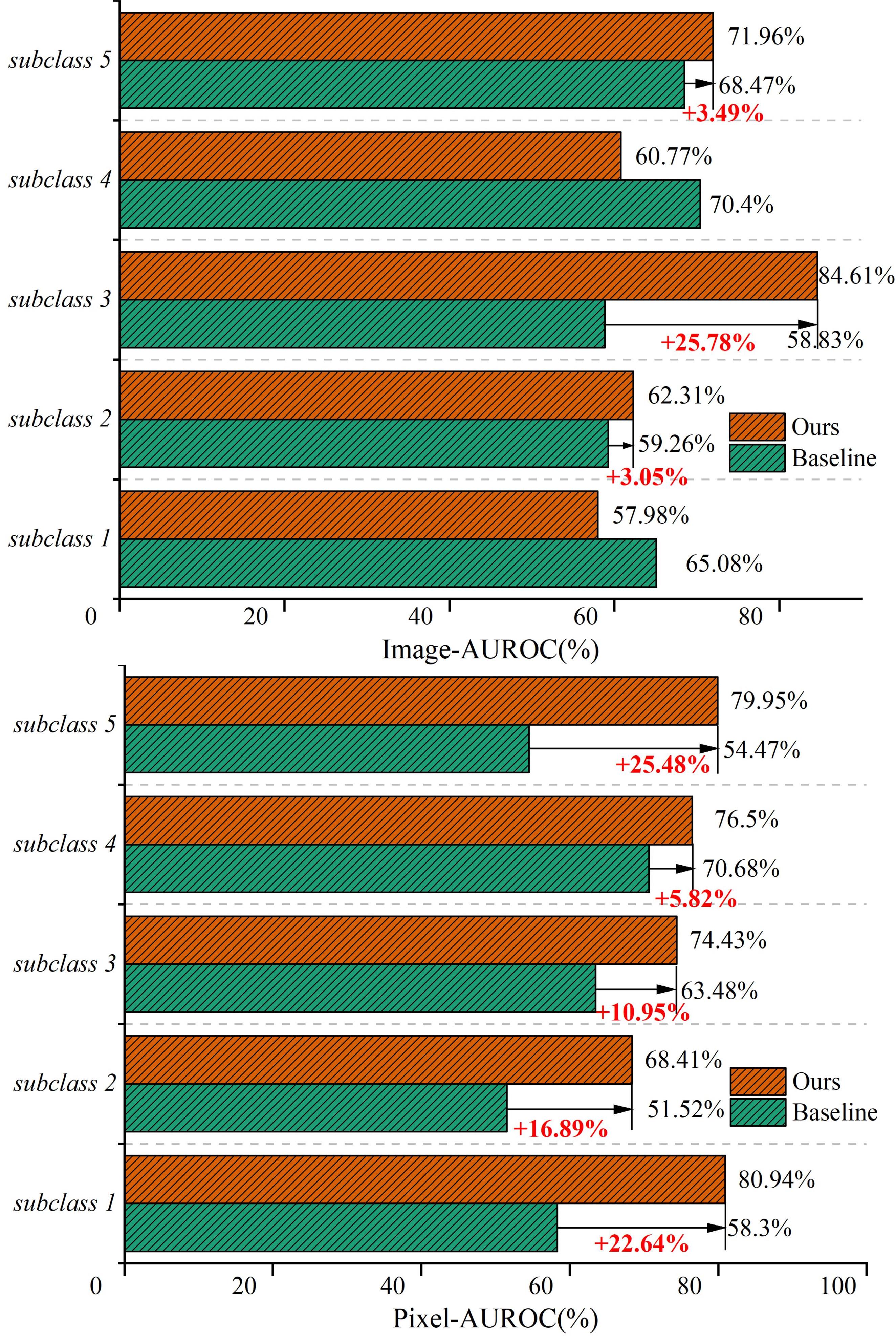} 
\caption{Ablation experiments for subclasses on cloth dataset. For the I-AUROC metric, our method improves on some unseen subclasses by $3.05\%$-$25.78\%$. For the P-AUROC metric, our method improves on all unseen subclasses by $5.82\%$-$25.48\%$.}
\label{figure:6}
\end{figure}
\begin{figure}[t]
\centering
\includegraphics[width=0.45\textwidth]{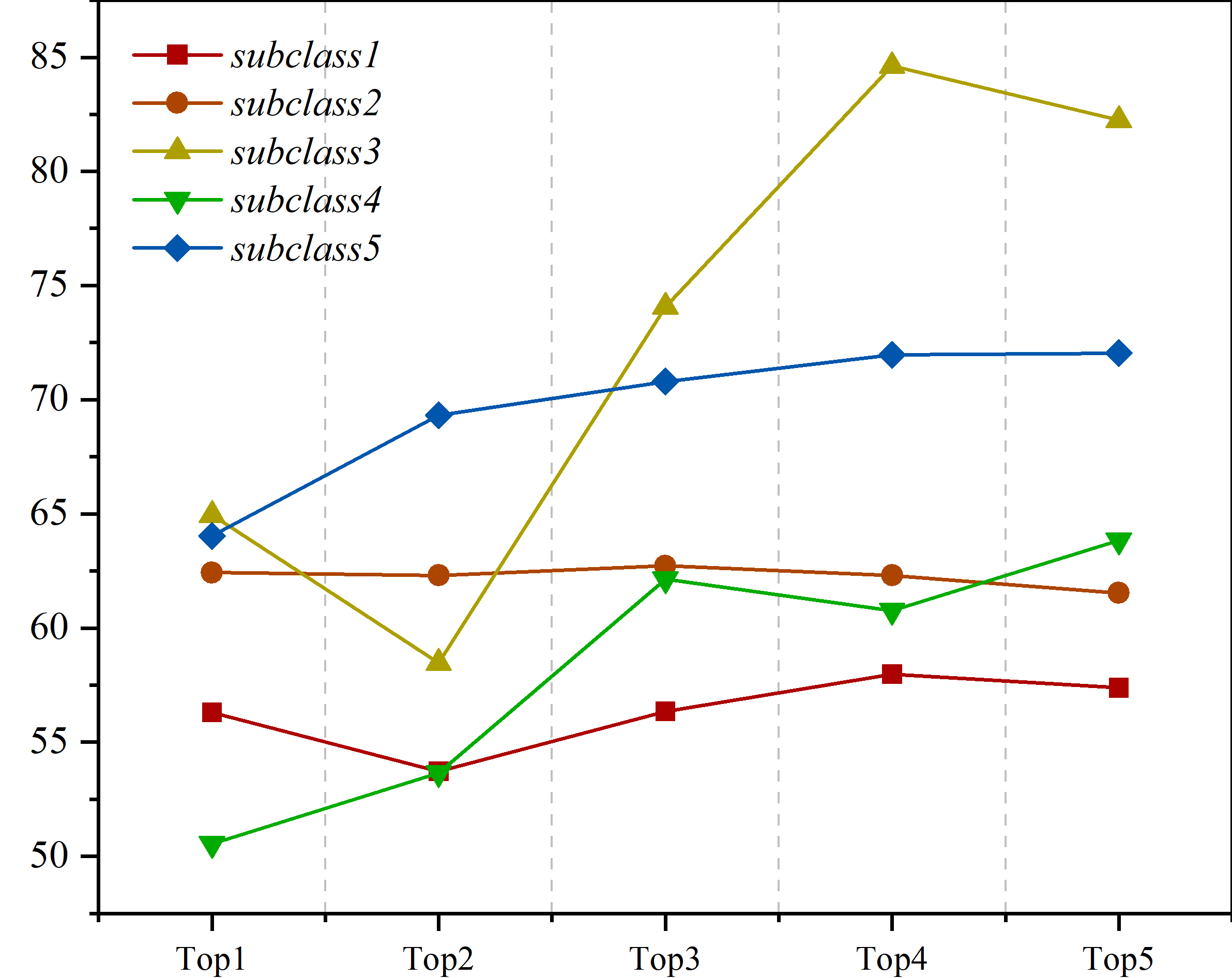} 
\caption{Ablation experiments for $Top k$ on cloth dataset. The results show that our method is optimal in choosing $Top 4$.}
\label{figure:7}
\end{figure}
\subsection{Ablation Studies}
In this section, we present ablation studies on the proposed method, including the structure of Adapted-MoE, the top $k$ number of MoE and the choice of the loss function in the routing network. The baseline for all ablation experiments in this section is SimpleNet.

\textbf{The Structure of Adapted-MoE.}
 To verify the validity of our proposed method, we conducted ablation experiments on cloth data, wafer data and metal data in the Texture AD dataset. As shown in Table \ref{tab:3}, using the MoE individually ignores the bias between the distribution of test samples and the distribution of samples that have been learned, leading to shortcomings in anomaly detection and anomaly location. In cloth data and wafer data, independent usage of Test-Time Adaption for feature embeddings can improve anomaly location performance, but the anomaly detection capability is greatly reduced due to the feature embeddings are not well assigned to the corresponding subclass space. Due to the small inter-class differences of the subclasses in the metal data(proved by visualization in the Appendix), MoE and normalization will lead to the wrong division of the subclasses into subspaces and only Test-Time Adaptation is needed to bring accuracy increase, $13.9\%/7.20\%$ of average P-AUROC and I-AUROC. Using both MoE and Test-Time Adaption will make the distribution of test samples not normalized correctly. Therefore, we introduced all the proposed methods into the baseline, which eventually improved $16.3\%/3.12\%$ and $1.57\%/2.18\%$ of average P-AUROC and I-AUROC on the cloth and wafer datasets, respectively. More details ablation results of subclasses can be found in the Appendix.

 Furthermore, we provide the ablation experimental results of the final proposed method for all subclasses in the cloth dataset. As shown in Figure \ref{figure:6}, the implementation of our proposed method improves the performance of anomaly detection in most of the subclasses by up to $25.48\%$. For anomaly location, our approach improves the performance of all subclasses with a maximum improvement of $25.48\%$ and a minimum improvement of $5.82\%$.

\textbf{The Top $k$ for Mixture of Experts.} The routing network identifies the expert model that is most closely associated with the test data, thereby minimizing the distance between the test samples. Furthermore, this approach can be employed to select the $Top k$ expert models that are most closely aligned with the test data. As shown in Figure \ref{figure:7}, we perform ablation experiments on the cloth dataset for $Top k$ expert model choices. The results show that in selecting $Top 4$ expert models is more beneficial to the overall model performance.
  
\textbf{Choice of Loss Function in Routing Network.} Due to the small scale of variation within the same category of data, the loss function determines for routing networks whether they can better distinguish between different subclasses. We compared the effect of softmax loss and center loss on the average performance of the three categories of datasets, as shown in Table \ref{tab:4}. The results show that center loss can better improve the performance of the routing network. This demonstrates that the addition of a centroid constraint can lead to a more explicit subclass space delineation.

\section{Conclusion}
    In this paper, we propose an Adapted-MoE for addressing the data variation and bias in the same category for anomaly detection. We define the issue of the variation of feature distribution within the training data in the real world leading to failure of the single decision boundary. Furthermore, we address the challenge of bias between the test and training data. We propose a Mixture of Experts that divides same-category samples into different feature spaces via a routing network, with each expert model constructing its own independent decision boundary. We use normalization to make the samples more uniformly distributed in the feature space. In addition, we propose a Test-Time Adaption to eliminate the bias between the distribution of test samples and learned features. Extensive experiments on Texture AD demonstrate that Adapted-MoE can be simply and efficiently implemented for anomaly detection and localization. 
    \subsubsection{Limitation.} This paper proposes a MoE for constructing multiple independent subclass decision boundaries. When using a dataset with a low diversity of subclasses, the performance improvement from MoE is lower than without MoE ($9.24\%\uparrow$ to $13.90\%\uparrow$) due to over-division being redundant. In addition, an overly complex expert model design will trigger overfitting in subclass learning. Therefore, the improvement effect is more limited to algorithms with a large number of parameters. In the future, we will focus on solving the overfitting problem caused by model complexity and data mismatch, aiming for greater improvements in more complex models \cite{you2022unified,zhou2023anomalyclip}.

\bibliography{aaai25}

\begin{thebibliography}{45}
\providecommand{\natexlab}[1]{#1}

\bibitem[{Ak{\c{c}}ay, Atapour-Abarghouei, and Breckon(2019)}]{akccay2019skip}
Ak{\c{c}}ay, S.; Atapour-Abarghouei, A.; and Breckon, T.~P. 2019.
\newblock Skip-ganomaly: Skip connected and adversarially trained encoder-decoder anomaly detection.
\newblock In \emph{2019 International Joint Conference on Neural Networks (IJCNN)}, 1--8. IEEE.

\bibitem[{Batzner, Heckler, and K{\"o}nig(2024)}]{batzner2024efficientad}
Batzner, K.; Heckler, L.; and K{\"o}nig, R. 2024.
\newblock Efficientad: Accurate visual anomaly detection at millisecond-level latencies.
\newblock In \emph{Proceedings of the IEEE/CVF Winter Conference on Applications of Computer Vision}, 128--138.

\bibitem[{Baur et~al.(2019)Baur, Wiestler, Albarqouni, and Navab}]{baur2019deep}
Baur, C.; Wiestler, B.; Albarqouni, S.; and Navab, N. 2019.
\newblock Deep autoencoding models for unsupervised anomaly segmentation in brain MR images.
\newblock In \emph{Brainlesion: Glioma, Multiple Sclerosis, Stroke and Traumatic Brain Injuries: 4th International Workshop, BrainLes 2018, Held in Conjunction with MICCAI 2018, Granada, Spain, September 16, 2018, Revised Selected Papers, Part I 4}, 161--169. Springer.

\bibitem[{Bergmann et~al.(2019{\natexlab{a}})Bergmann, Fauser, Sattlegger, and Steger}]{bergmann2019mvtec}
Bergmann, P.; Fauser, M.; Sattlegger, D.; and Steger, C. 2019{\natexlab{a}}.
\newblock MVTec AD--A comprehensive real-world dataset for unsupervised anomaly detection.
\newblock In \emph{Proceedings of the IEEE/CVF conference on computer vision and pattern recognition}, 9592--9600.

\bibitem[{Bergmann et~al.(2019{\natexlab{b}})Bergmann, L{\"o}we, Fauser, Sattlegger, and Steger}]{bergmann2019improving}
Bergmann, P.; L{\"o}we, S.; Fauser, M.; Sattlegger, D.; and Steger, C. 2019{\natexlab{b}}.
\newblock Improving Unsupervised Defect Segmentation by Applying Structural Similarity to Autoencoders.
\newblock In \emph{Proceedings of the 14th International Joint Conference on Computer Vision, Imaging and Computer Graphics Theory and Applications}. SCITEPRESS-Science and Technology Publications.

\bibitem[{Cao, Zhu, and Pang(2023)}]{cao2023anomaly}
Cao, T.; Zhu, J.; and Pang, G. 2023.
\newblock Anomaly detection under distribution shift.
\newblock In \emph{Proceedings of the IEEE/CVF International Conference on Computer Vision}, 6511--6523.

\bibitem[{Cao et~al.(2024)Cao, Xu, Zhang, Cheng, Huang, Pang, and Shen}]{cao2024survey}
Cao, Y.; Xu, X.; Zhang, J.; Cheng, Y.; Huang, X.; Pang, G.; and Shen, W. 2024.
\newblock A survey on visual anomaly detection: Challenge, approach, and prospect.
\newblock \emph{arXiv preprint arXiv:2401.16402}.

\bibitem[{Contreras-Cruz et~al.(2023)Contreras-Cruz, Correa-Tome, Lopez-Padilla, and Ramirez-Paredes}]{contreras2023generative}
Contreras-Cruz, M.~A.; Correa-Tome, F.~E.; Lopez-Padilla, R.; and Ramirez-Paredes, J.-P. 2023.
\newblock Generative Adversarial Networks for anomaly detection in aerial images.
\newblock \emph{Computers and Electrical Engineering}, 106: 108470.

\bibitem[{Dai et~al.(2024)Dai, Wu, Li, and Xue}]{dai2024generating}
Dai, S.; Wu, Y.; Li, X.; and Xue, X. 2024.
\newblock Generating and reweighting dense contrastive patterns for unsupervised anomaly detection.
\newblock In \emph{Proceedings of the AAAI Conference on Artificial Intelligence}, volume~38, 1454--1462.

\bibitem[{Deng et~al.(2009)Deng, Dong, Socher, Li, Li, and Fei-Fei}]{deng2009imagenet}
Deng, J.; Dong, W.; Socher, R.; Li, L.-J.; Li, K.; and Fei-Fei, L. 2009.
\newblock Imagenet: A large-scale hierarchical image database.
\newblock In \emph{2009 IEEE conference on computer vision and pattern recognition}, 248--255. Ieee.

\bibitem[{Duan et~al.(2023)Duan, Hong, Niu, and Zhang}]{duan2023few}
Duan, Y.; Hong, Y.; Niu, L.; and Zhang, L. 2023.
\newblock Few-shot defect image generation via defect-aware feature manipulation.
\newblock In \emph{Proceedings of the AAAI Conference on Artificial Intelligence}, volume~37, 571--578.

\bibitem[{Goodfellow et~al.(2014)Goodfellow, Pouget-Abadie, Mirza, Xu, Warde-Farley, Ozair, Courville, and Bengio}]{goodfellow2014generative}
Goodfellow, I.; Pouget-Abadie, J.; Mirza, M.; Xu, B.; Warde-Farley, D.; Ozair, S.; Courville, A.; and Bengio, Y. 2014.
\newblock Generative adversarial nets.
\newblock \emph{Advances in neural information processing systems}, 27.

\bibitem[{Haselmann, Gruber, and Tabatabai(2018)}]{haselmann2018anomaly}
Haselmann, M.; Gruber, D.~P.; and Tabatabai, P. 2018.
\newblock Anomaly detection using deep learning based image completion.
\newblock In \emph{2018 17th IEEE international conference on machine learning and applications (ICMLA)}, 1237--1242. IEEE.

\bibitem[{Heckler, K{\"o}nig, and Bergmann(2023)}]{heckler2023exploring}
Heckler, L.; K{\"o}nig, R.; and Bergmann, P. 2023.
\newblock Exploring the importance of pretrained feature extractors for unsupervised anomaly detection and localization.
\newblock In \emph{Proceedings of the IEEE/CVF Conference on Computer Vision and Pattern Recognition}, 2917--2926.

\bibitem[{Hu et~al.(2024)Hu, Chen, Gan, Peng, Zhang, Zhang, Wang, Wang, Cao, and Ji}]{hu2024dmad}
Hu, J.; Chen, X.; Gan, Z.; Peng, J.; Zhang, S.; Zhang, J.; Wang, Y.; Wang, C.; Cao, L.; and Ji, R. 2024.
\newblock DMAD: Dual Memory Bank for Real-World Anomaly Detection.
\newblock \emph{arXiv preprint arXiv:2403.12362}.

\bibitem[{Lei et~al.(2023)Lei, Hu, Wang, and Liu}]{lei2023pyramidflow}
Lei, J.; Hu, X.; Wang, Y.; and Liu, D. 2023.
\newblock Pyramidflow: High-resolution defect contrastive localization using pyramid normalizing flow.
\newblock In \emph{Proceedings of the IEEE/CVF conference on computer vision and pattern recognition}, 14143--14152.

\bibitem[{Li, Zhu, and Van~Leeuwen(2023)}]{li2023survey}
Li, Z.; Zhu, Y.; and Van~Leeuwen, M. 2023.
\newblock A survey on explainable anomaly detection.
\newblock \emph{ACM Transactions on Knowledge Discovery from Data}, 18(1): 1--54.

\bibitem[{Liang et~al.(2023)Liang, Zhang, Zhao, Wu, Liu, and Pan}]{liang2023omni}
Liang, Y.; Zhang, J.; Zhao, S.; Wu, R.; Liu, Y.; and Pan, S. 2023.
\newblock Omni-frequency channel-selection representations for unsupervised anomaly detection.
\newblock \emph{IEEE Transactions on Image Processing}.

\bibitem[{Lin and Yan(2024)}]{lin2024comprehensive}
Lin, J.; and Yan, Y. 2024.
\newblock A Comprehensive Augmentation Framework for Anomaly Detection.
\newblock In \emph{Proceedings of the AAAI Conference on Artificial Intelligence}, volume~38, 8742--8749.

\bibitem[{Liu, Tan, and Zhou(2022)}]{liu2022unsupervised}
Liu, B.; Tan, P.-N.; and Zhou, J. 2022.
\newblock Unsupervised anomaly detection by robust density estimation.
\newblock In \emph{Proceedings of the AAAI Conference on Artificial Intelligence}, volume~36, 4101--4108.

\bibitem[{Liu et~al.(2024)Liu, Xie, Wang, Li, Wang, Zheng, and Jin}]{liu2024deep}
Liu, J.; Xie, G.; Wang, J.; Li, S.; Wang, C.; Zheng, F.; and Jin, Y. 2024.
\newblock Deep industrial image anomaly detection: A survey.
\newblock \emph{Machine Intelligence Research}, 21(1): 104--135.

\bibitem[{Liu et~al.(2023)Liu, Zhou, Xu, and Wang}]{liu2023simplenet}
Liu, Z.; Zhou, Y.; Xu, Y.; and Wang, Z. 2023.
\newblock Simplenet: A simple network for image anomaly detection and localization.
\newblock In \emph{Proceedings of the IEEE/CVF Conference on Computer Vision and Pattern Recognition}, 20402--20411.

\bibitem[{Mishra et~al.(2021)Mishra, Verk, Fornasier, Piciarelli, and Foresti}]{mishra2021vt}
Mishra, P.; Verk, R.; Fornasier, D.; Piciarelli, C.; and Foresti, G.~L. 2021.
\newblock VT-ADL: A vision transformer network for image anomaly detection and localization.
\newblock In \emph{2021 IEEE 30th International Symposium on Industrial Electronics (ISIE)}, 01--06. IEEE.

\bibitem[{Mousakhan, Brox, and Tayyub(2023)}]{mousakhan2023anomaly}
Mousakhan, A.; Brox, T.; and Tayyub, J. 2023.
\newblock Anomaly detection with conditioned denoising diffusion models.
\newblock \emph{arXiv preprint arXiv:2305.15956}.

\bibitem[{Park, Noh, and Ham(2020)}]{park2020learning}
Park, H.; Noh, J.; and Ham, B. 2020.
\newblock Learning memory-guided normality for anomaly detection.
\newblock In \emph{Proceedings of the IEEE/CVF conference on computer vision and pattern recognition}, 14372--14381.

\bibitem[{Reiss and Hoshen(2023)}]{reiss2023mean}
Reiss, T.; and Hoshen, Y. 2023.
\newblock Mean-shifted contrastive loss for anomaly detection.
\newblock In \emph{Proceedings of the AAAI Conference on Artificial Intelligence}, volume~37, 2155--2162.

\bibitem[{Ristea et~al.(2022)Ristea, Madan, Ionescu, Nasrollahi, Khan, Moeslund, and Shah}]{ristea2022self}
Ristea, N.-C.; Madan, N.; Ionescu, R.~T.; Nasrollahi, K.; Khan, F.~S.; Moeslund, T.~B.; and Shah, M. 2022.
\newblock Self-supervised predictive convolutional attentive block for anomaly detection.
\newblock In \emph{Proceedings of the IEEE/CVF conference on computer vision and pattern recognition}, 13576--13586.

\bibitem[{Roth et~al.(2022)Roth, Pemula, Zepeda, Sch{\"o}lkopf, Brox, and Gehler}]{roth2022towards}
Roth, K.; Pemula, L.; Zepeda, J.; Sch{\"o}lkopf, B.; Brox, T.; and Gehler, P. 2022.
\newblock Towards total recall in industrial anomaly detection.
\newblock In \emph{Proceedings of the IEEE/CVF conference on computer vision and pattern recognition}, 14318--14328.

\bibitem[{Sabokrou et~al.(2018)Sabokrou, Khalooei, Fathy, and Adeli}]{sabokrou2018adversarially}
Sabokrou, M.; Khalooei, M.; Fathy, M.; and Adeli, E. 2018.
\newblock Adversarially learned one-class classifier for novelty detection.
\newblock In \emph{Proceedings of the IEEE conference on computer vision and pattern recognition}, 3379--3388.

\bibitem[{Schlegl et~al.(2017)Schlegl, Seeb{\"o}ck, Waldstein, Schmidt-Erfurth, and Langs}]{schlegl2017unsupervised}
Schlegl, T.; Seeb{\"o}ck, P.; Waldstein, S.~M.; Schmidt-Erfurth, U.; and Langs, G. 2017.
\newblock Unsupervised anomaly detection with generative adversarial networks to guide marker discovery.
\newblock In \emph{International conference on information processing in medical imaging}, 146--157. Springer.

\bibitem[{Texture-ad(2024)}]{texturead}
Texture-ad. 2024.
\newblock Texture-AD-Benchmark.
\newblock \url{https://huggingface.co/datasets/texture-ad/Texture-AD-Benchmark}.
\newblock Accessed: 2024-08-15.

\bibitem[{Wang et~al.(2023)Wang, Peng, Zhang, Yi, Wang, and Wang}]{wang2023multimodal}
Wang, Y.; Peng, J.; Zhang, J.; Yi, R.; Wang, Y.; and Wang, C. 2023.
\newblock Multimodal industrial anomaly detection via hybrid fusion.
\newblock In \emph{Proceedings of the IEEE/CVF Conference on Computer Vision and Pattern Recognition}, 8032--8041.

\bibitem[{Wen et~al.(2016)Wen, Zhang, Li, and Qiao}]{wen2016discriminative}
Wen, Y.; Zhang, K.; Li, Z.; and Qiao, Y. 2016.
\newblock A discriminative feature learning approach for deep face recognition.
\newblock In \emph{Computer vision--ECCV 2016: 14th European conference, amsterdam, the netherlands, October 11--14, 2016, proceedings, part VII 14}, 499--515. Springer.

\bibitem[{Wu et~al.(2024)Wu, Fan, Zhou, Yu, Deng, Zou, and Lin}]{wu2024unsupervised}
Wu, D.; Fan, S.; Zhou, X.; Yu, L.; Deng, Y.; Zou, J.; and Lin, B. 2024.
\newblock Unsupervised Anomaly Detection via Masked Diffusion Posterior Sampling.
\newblock \emph{arXiv preprint arXiv:2404.17900}.

\bibitem[{Yan et~al.(2021)Yan, Zhang, Xu, Hu, and Heng}]{yan2021learning}
Yan, X.; Zhang, H.; Xu, X.; Hu, X.; and Heng, P.-A. 2021.
\newblock Learning semantic context from normal samples for unsupervised anomaly detection.
\newblock In \emph{Proceedings of the AAAI conference on artificial intelligence}, volume~35, 3110--3118.

\bibitem[{Yang et~al.(2023)Yang, Liu, Yang, and Wu}]{yang2023slsg}
Yang, M.; Liu, J.; Yang, Z.; and Wu, Z. 2023.
\newblock Slsg: Industrial image anomaly detection by learning better feature embeddings and one-class classification.
\newblock \emph{arXiv preprint arXiv:2305.00398}.

\bibitem[{You et~al.(2022)You, Cui, Shen, Yang, Lu, Zheng, and Le}]{you2022unified}
You, Z.; Cui, L.; Shen, Y.; Yang, K.; Lu, X.; Zheng, Y.; and Le, X. 2022.
\newblock A unified model for multi-class anomaly detection.
\newblock \emph{Advances in Neural Information Processing Systems}, 35: 4571--4584.

\bibitem[{Zagoruyko and Komodakis(2016)}]{zagoruyko2016wide}
Zagoruyko, S.; and Komodakis, N. 2016.
\newblock Wide residual networks.
\newblock \emph{arXiv preprint arXiv:1605.07146}.

\bibitem[{Zavrtanik, Kristan, and Sko{\v{c}}aj(2021{\natexlab{a}})}]{zavrtanik2021draem}
Zavrtanik, V.; Kristan, M.; and Sko{\v{c}}aj, D. 2021{\natexlab{a}}.
\newblock Draem-a discriminatively trained reconstruction embedding for surface anomaly detection.
\newblock In \emph{Proceedings of the IEEE/CVF international conference on computer vision}, 8330--8339.

\bibitem[{Zavrtanik, Kristan, and Sko{\v{c}}aj(2021{\natexlab{b}})}]{zavrtanik2021reconstruction}
Zavrtanik, V.; Kristan, M.; and Sko{\v{c}}aj, D. 2021{\natexlab{b}}.
\newblock Reconstruction by inpainting for visual anomaly detection.
\newblock \emph{Pattern Recognition}, 112: 107706.

\bibitem[{Zhang et~al.(2023{\natexlab{a}})Zhang, Wang, Wu, and Jiang}]{zhang2023diffusionad}
Zhang, H.; Wang, Z.; Wu, Z.; and Jiang, Y.-G. 2023{\natexlab{a}}.
\newblock Diffusionad: Denoising diffusion for anomaly detection.
\newblock \emph{arXiv preprint arXiv:2303.08730}, 4.

\bibitem[{Zhang et~al.(2023{\natexlab{b}})Zhang, Li, Li, Huang, Shan, and Chen}]{zhang2023destseg}
Zhang, X.; Li, S.; Li, X.; Huang, P.; Shan, J.; and Chen, T. 2023{\natexlab{b}}.
\newblock Destseg: Segmentation guided denoising student-teacher for anomaly detection.
\newblock In \emph{Proceedings of the IEEE/CVF Conference on Computer Vision and Pattern Recognition}, 3914--3923.

\bibitem[{Zhou et~al.(2020)Zhou, Xiao, Yang, Cheng, Liu, Luo, Gu, Liu, and Gao}]{zhou2020encoding}
Zhou, K.; Xiao, Y.; Yang, J.; Cheng, J.; Liu, W.; Luo, W.; Gu, Z.; Liu, J.; and Gao, S. 2020.
\newblock Encoding structure-texture relation with p-net for anomaly detection in retinal images.
\newblock In \emph{Computer Vision--ECCV 2020: 16th European Conference, Glasgow, UK, August 23--28, 2020, Proceedings, Part XX 16}, 360--377. Springer.

\bibitem[{Zhou et~al.(2023)Zhou, Pang, Tian, He, and Chen}]{zhou2023anomalyclip}
Zhou, Q.; Pang, G.; Tian, Y.; He, S.; and Chen, J. 2023.
\newblock Anomalyclip: Object-agnostic prompt learning for zero-shot anomaly detection.
\newblock \emph{arXiv preprint arXiv:2310.18961}.

\bibitem[{Zhou et~al.(2024)Zhou, Xu, Song, Shen, and Shen}]{zhou2024msflow}
Zhou, Y.; Xu, X.; Song, J.; Shen, F.; and Shen, H.~T. 2024.
\newblock Msflow: Multiscale flow-based framework for unsupervised anomaly detection.
\newblock \emph{IEEE Transactions on Neural Networks and Learning Systems}.

\end{thebibliography}

\end{document}